\documentclass[opre, nonblindrev]{informs3} 
\OneAndAHalfSpacedXI 

\usepackage{amsmath}
\usepackage{amssymb}
\usepackage{comment}
\usepackage{subfigure}
\usepackage{bm}
\usepackage{url}
\usepackage{graphicx}
\usepackage{epstopdf}
\usepackage{epsfig,psfrag}
\usepackage{verbatim} 
\usepackage{subfigure}
\usepackage{algorithmicx,algpseudocode}
\usepackage{algorithm,float}
\makeatletter
\renewcommand{\ALG@name}{Strategy:}

\makeatother
\usepackage{enumerate}
\usepackage[dvipsnames]{xcolor}

\usepackage{pdfsync}
\usepackage[normalem]{ulem}

\def\N{\mathbb{N}}
\def\R{\mathbb{R}}

\def\pb{\mathbb{P}}

\def\calE{\mathcal{E}}

\def\calG{\mathcal{G}}

\def\calJ{\mathcal{J}}
\def\calI{\mathcal{I}}

\def\calQ{\mathcal{Q}}
\def\calX{\mathcal{X}}
\def\calY{\mathcal{Y}}

\def\calX{\mathcal{X}}

\def\what{\widehat}

\def\l2#1{\left\|#1\right\|_2}

\def\bpf{\proof{Proof.}}

\newcommand{\lt}{\left}
\newcommand{\rt}{\right}

\theoremstyle{plain}
\newtheorem{theorem}{Theorem}[section]
	
\newtheorem{lemma}[theorem]{Lemma}

\newtheorem{proposition}[theorem]{Proposition}

\newtheorem{corollary}[theorem]{Corollary}
\newtheorem{definition}[theorem]{Definition}

\newtheorem{claim}[theorem]{Claim}

\usepackage{natbib}
 \bibpunct[, ]{(}{)}{,}{a}{}{,}%
 \def\bibfont{\small}%
\EquationsNumberedThrough

\begin{document}



\RUNAUTHOR{Tsitsiklis, Xu, and Xu}
\RUNTITLE{Private Sequential Learning}

\TITLE{Private Sequential Learning}


\ARTICLEAUTHORS{%
John N. Tsitsiklis
\AFF{LIDS, Massachusetts Institute of Technology, Cambridge, MA 02139, \EMAIL{jnt@mit.edu}} 
Kuang Xu
\AFF{Graduate School of Business, Stanford University, Stanford, CA 94305, \EMAIL{kuangxu@stanford.edu}}
Zhi Xu
\AFF{LIDS, Massachusetts Institute of Technology, Cambridge, MA 02139, \EMAIL{zhixu@mit.edu}} 

}  

\ABSTRACT{

{We {formulate} a private {learning} model to study an intrinsic tradeoff between privacy and query complexity in {sequential learning}. Our model involves a {learner} who aims to determine a scalar value, $v^*$, {by} sequentially querying an external database and receiving binary responses. {In the meantime, an adversary observes the {learner}'s queries, though not 
{the} responses, and tries to infer from them the value of $v^*$.} {The objective of the {learner} is to obtain an accurate estimate of $v^*$ using only a small number of queries, while simultaneously protecting her privacy by making $v^*$ provably difficult to learn for the adversary.}} {Our main results} provide tight upper and lower bounds on the {learner}'s {query complexity} as a function of desired levels of privacy and {estimation accuracy}. We also construct explicit query strategies whose complexity is optimal up to an additive constant.
\footnote{Conference: an extended abstract of this paper appeared in the proceedings of the Conference on Learning Theory (COLT) 2018. Journal: accepted at Operations Research, 2020. }


}

\maketitle

\section{Introduction}

Organizations and individuals often rely on relevant data to solve decision problems. Sometimes, such data are beyond the immediate reach of a decision maker and must be acquired 
{by}
interacting with an external entity or environment. However, these interactions may be monitored by a third-party adversary and subject the decision maker to potential privacy breaches, a possibility that has become increasingly prominent as information technologies and tools for data analytics advance. 

The present paper studies a decision maker, henceforth referred to as the {{learner}}, who acquires data from an external entity in an {interactive} fashion by submitting sequential queries. The \emph{interactivity} benefits the {learner} by enabling her to tailor future queries based on past responses and thus reduce the number of queries needed, while, at the same time, exposes the {learner} to substantial privacy risk: the more her queries depend on past responses, the easier it might be for an {adversary} to use the observed queries to infer those past responses. 
Our main objective is to articulate and understand an intrinsic \emph{privacy versus query complexity tradeoff} in the context of such a {Private Sequential Learning} model. 

We begin with an informal description of {the} model. A \emph{{learner}} would like to determine the value of a scalar, $v^*$, referred to as the \emph{true value}, which lies in a bounded subset of $\mathbb{R}${, for example, the interval [0,1)}. To search for $v^*$, she must interact with an external database, through sequentially submitted queries: at step $k$, the {learner} submits a query, $q_k \in \R$, and receives a binary response, $r_k$, where $r_k=1$ if $v^*\geq q_k$, and $r_k=0$, otherwise. The interaction is sequential in the sense that the {learner} may choose a query  depending on the responses to all previous queries. {Meanwhile}, there is an \emph{adversary} who eavesdrops on the {learner}'s actions: she observes all of the {learner}'s queries, $q_k$, but not 
{the} 
responses, and tries to use these queries to estimate the true value, $v^*$. The {learner}'s goal is to submit queries in such a way that she can learn $v^*$ within a prescribed error tolerance, while $v^*$ cannot be accurately estimated by the adversary with high confidence. {The learner's goal is easily attained by submitting} an unlimited number of queries, in which case the queries need not depend on the past responses and hence reveal no information to the adversary. Our quest is, however, to understand the \emph{least number of queries} that the {learner} needs to submit in order for her to successfully retain privacy. Is the query complexity significantly different from the case where privacy constraints are absent? How does it vary as a function of the levels of accuracy and  privacy? Is there a simple and yet efficient query strategy that the {learner} can adopt?  Our main results address these questions.

\subsection{Motivating Examples}\label{sec:motivating_examples}
We discuss {three} examples that provide some context for our model. {While the examples are stylized, they are intended to motivate and provide insights on the general concept of privacy-preserving in sequential learning.} 

\emph{Example 1 - learning an optimal price.} A firm is to release a new product and would like to identify a revenue maximizing price, $p^*$, prior to the product launch. 
{
The firm believes that there is an underlying unknown demand function, $g(p)$, and that the revenue function $f(p) = g(p)*p$ is  a strictly concave and differentiable function of the price, $p$.}
A sequential {learning} process is employed to identify $p^*$ over a series of epochs: in epoch $k$, the firm assesses how the market responds to a test price, $p_k$, and receives a binary feedback as to whether $f'(p_k)\geq0$ or  $f'(p_k)< 0$. This may be achieved, for instance, by contracting a consulting firm to conduct market surveys on the price {\emph{sensitivity}} around $p_k$. {The survey could ask for a customer's willingness to purchase the product at price levels $p$ and $p+\xi$, for some small $\xi$, from which the demand $g(p)$ and its gradient can be estimated. Using the chain rule, this information can then be converted to an {estimate of} the sign of $f'(p)$.} The firm would like to {be able to} estimate $p^*$ with reasonable accuracy {after} a small number of epochs, but is wary that a competitor might be able to observe the surveys, {{by} either purposely participat{ing} in {them} or {by} interview{ing} {survey participants,} }and {then} deduce  the value of $p^*$ ahead of the product launch. In the context of Private Sequential {Learning}, the firm is the {learner}, the competitor is the adversary, the revenue-maximizing price is the true value, and the test prices are the queries. The binary response on the revenue's price sensitivity indicates whether the revenue-maximizing price is less than the current test price. 

{\emph{Example 2 - learning consumer preference{s}.} E-commence firms, like Amazon, are often incentivized to learn consumer preferences that could later be used for devising personalized promotions. Consider a consumer who is looking for an item with an ideal scalar product feature (e.g., size) on an online merchant's platform. While the consumer does not initially know the optimal feature value, when presented with a specific product, she will be able to assess whether the ideal value is greater than the current option (e.g., if the current option is too large or too small). The consumer browses different products in a sequential manner, and hopes to eventually narrow in on the ideal feature value. Meanwhile, the platform sees all of the products {browsed} by the consumer, but does not directly observe her internal assessments. Our model can be used to investigate how a privacy-aware consumer can browse products in such a manner that will eventually allow her to identify the ideal product feature value, while preventing the platform from confidently inferring 
{that} value from her browsing history.}

\emph{Example {3} - online optimization with private weights.}  In the previous examples, the adversary is a third-party that does not observe the responses to the queries. We now provide a different example in which the adversary is the database to which queries are submitted, and thus has partial knowledge of the responses.  Consider a {learner} who wishes to identify the maximizer, $x^*$, of a function $f(x) = \sum_{i=1}^m \alpha_i f_i(x)$ over some bounded interval $\calX \subset \R$, where $\{f_i(\cdot)\}_{1\leq i \leq m}$ is a collection of strictly concave differentiable constituent functions, and $\{\alpha_i\}_{1\leq i \leq m}$ are positive (private) weights representing the importance that the {learner} associates with each constituent function. The {learner} knows the weights but does not have information about the constituent functions; such knowledge is to be acquired by querying an external database. During epoch $k$, the {learner} submits a test value, $x_k$, and receives from the database the derivatives of all constituent functions at $x_k$, $\{f'_i(x_k)\}_{1\leq i \leq m}$. Using the weights, the {learner} can then compute the derivative $f'(x_k)$, whose sign serves as a binary indicator of the position of the maximizer $x^*$ relative to the current test value. The database, which possesses complete information about the constituent functions but does not know the weights, would like to infer from the {learner}'s querying pattern the maximizing value $x^*$ or possibly the weights themselves. {While the above-mentioned model appears to differ from the {previous two} examples, it turns out that the modeling methodology and query strategies that we develop 
 can also be applied to this setting. The connection between the two {settings} is made precise in Chapter 2 of \cite{zhi}.}

\subsection{Preview of the Main Result}

We now 
{preview} our main result. Let us begin by introducing some additional notation. Recall that both the {learner} and the adversary aim to obtain estimates that are close to a true value {$v^*\in[0,1)$}. We denote by {$\epsilon{/2}$ and $\delta{/2}$ the {estimation error}} that the {learner} and the adversary is willing to tolerate, respectively. {We will employ a privacy parameter $L\in\mathbb{N}$ to quantify the {learner}'s level of privacy at the end of the {learning} process: the {learner}'s privacy level is $L$ if the adversary {can} successfully approximate the true value within an  error of {$\delta/2$} with probability at most $1/L$}, {so that higher values of $L$ correspond to enhanced privacy.} 
{(A  precise, formal definition of ``privacy level,'' in terms of $L$, will be provided in Section~\ref{sec:stronger_privacy}.)}
A private query strategy for the {learner} {then} must be able to produce an estimate of the {true value within an error of at most} {$\epsilon/{2}$}, while simultaneously guaranteeing that the desired privacy level $L$ holds against the adversary.  

{Our main objective is to quantify the \emph{query complexity} of private sequential learning, $N^*(\epsilon,\delta,L)$, defined as the minimum number of queries needed for a private {learner} strategy, under a given set of parameters, $\epsilon, \delta${,} and $L$. {Specifically, we} will focus on the regime where $2\epsilon<\delta\leq 1/L$. The reason 
{for} this choice will become clear after a formal introduction of the model, and we will revisit it at the beginning of Section \ref{sec:complex_results}.
In this regime, we have the following upper and lower bounds on the query complexity.

\begin{enumerate}
\item We establish an upper bound\footnote{All logarithms are taken with respect to base 2. To reduce clutter, non-integer numbers are to be understood as  rounded upwards. 
For example, the lower bound should be understood as $\lceil\log(1/L\epsilon)\rceil+2L$, where $\lceil\cdot\rceil$ represents the ceiling function.
} of $\log(1/L\epsilon)+2L$ by explicitly constructing a private {learner} strategy, which applies for any $\delta$ in the range $(2\epsilon, 1/L]$. 
	\item We establish a lower bound of $\log(\delta/\epsilon)+2L-4$ by characterizing the amount of  information available to the adversary.
\end{enumerate}
{We} {note} that {our bounds are tight in the sense that when the adversary's accuracy requirement is as loose as possible, i.e., $\delta=1/L$}, the upper bound {matches the lower bound, up to} an additive constant,  {equal to~4.}
{Furthermore, comparing with the elementary lower bound of $\log(1/\epsilon)$ for the case where privacy is not a concern, we see that the extra effort necessary to guarantee a privacy level $L$ is at most an additive factor of $2L$.}

{To put our results in context, we examine in Section \ref{sec:example_strategy} two simple strateg{ies} situated at two extreme points of the privacy-efficiency tradeoff curve. On the one hand, the classical bisection search algorithm achieves an optimal query complexity of $\log(1/\epsilon)$, {but} it completely reveals the responses of past queries and is hence almost never private. On the other hand, the learner could use a non-adaptive {``$\epsilon$-dense''} strategy {that places} $1/\epsilon$ equally spaced queries throughout the unit interval. The non-adaptive nature of this strategy {allows} the learner to be always private, {but} its query complexity is significantly worse. Our results and policies essentially aim to {understand} the optimal tradeoff between these two extremes, for any given privacy level, $L$. 
}

\subsection{Related Work}

In the absence of a privacy constraint, the problem of identifying a value within a compact interval through (possibly noisy)  binary feedback is a classical problem arising in domains such as coding theory (\cite{horstein1963sequential}) and root finding (\cite{waeber2013bisection}). It is well known that the bisection algorithm achieves the optimal query complexity of $\log(1/
{\epsilon})$ (cf.~\cite{waeber2013bisection}),
{where $\epsilon>0$ is the error tolerance.} In contrast, to the best of our knowledge, the question of how to preserve a {learner}'s privacy when her actions are fully observed by an adversary and what the resulting query complexity would be has received relatively little attention in the literature. 

{Related to our work, in spirit, is the body of literature on differential privacy (\cite{dwork2006calibrating,dwork2014algorithmic}), {a concept {that} has been applied in statistics (\cite{wasserman2010statistical,smith2011privacy,duchi2016minimax}) and learning theory (\cite{raskhodnikova2008can,chaudhuri2011sample,blum2013learning,feldman2014sample}).}  Differential privacy mandates that the output distribution of an algorithm be insensitive under 
{certain perturbations}
 of the input data. For instance, \cite{jain2012differentially} 
{study}
regret minimization in an online optimization problem while ensuring differential privacy, 
{in the sense that} the distribution of the sequence of solutions remains nearly identical when any {one} of the functions being optimized is perturbed.
{In \cite{dwork2012privacyanalyst}, the authors study differential privacy in data analysis when multiple analysts query the same database. More recently, adaptive versions of differentially private data analysis have been studied \citep{dwork2015generalization,dwork2015reusable,dwork2015preserving,cummings2016adaptive}. Notably, our work departs from this literature by using a goal-oriented privacy framework: our definition of privacy measures the adversary's ability to perform a \emph{specific} inferential goal.  In contrast, differential privacy aims to prevent an adversary from performing \emph{any} meaningful inference. As such, the goal-oriented privacy framework leads to substantially more efficient decision policies, while differential privacy offers stronger privacy protection for settings where the privacy goal is not clear \emph{a prior{i}}, at the cost of an increased efficiency loss. In this regard, our approach echoes a number of recent papers that also employ a goal-oriented privacy framework \citep{fanti2015spy,kuang, liao2018hypothesis}.}

In a different model, \cite{kuang} {study} the issue of privacy in a sequential decision problem, where an agent attempts to reach a particular node in a graph, {traversing it} in a way that obfuscates her intended destination{,} against an adversary who observes her past trajectory. The authors show that the probability of a correct prediction by the adversary is inversely proportional to the time it takes for the {agent} to reach her destination. Similar to the setting of \cite{kuang}, the {learner} in our model also plays against a powerful adversary who observes all past actions. However, a major new element  is that the {learner} in our model strives to \emph{learn} a piece of information of which she herself has no prior knowledge, in contrast to the {agent} in \cite{kuang} who tries to conceal private information already in her possession. In a way, the central conflict of trying to learn something while preventing others from learning the same information sets {the present} work apart from the extant literature.

{Finally, our model  is }
{close in spirit} to private information retrieval problems in  the field of cryptography (\cite{kushilevitz1997replication,chor1998private,gasarch2004survey}).  In these problems, a {learner} wishes to retrieve an item from some location $i$ in a database, in such a manner that the database obtains no information on the value of $i$, where the latter requirement can be either information{-}theoretic or based on computational hardness assumptions. Compared to this line of literature, our privacy requirement is substantially weaker: the adversary  may still obtain \emph{some} information on the true value. This relaxation of the privacy requirement allows the {learner} to deploy richer and more sample-efficient query strategies.

\subsection{Organization}
The remainder of the paper is organized as follows. We formally introduce the {Private Sequential {Learning} model in Section \ref{sec:model}. In Section \ref{sec:privatestrategycomplex}, we motivate and discuss private {learner} strategies. Our main results are stated in Section \ref{sec:complex_results}. Before delving into the proofs, we examine in Section \ref{sec:example_strategy} three  examples of {learner} strategies that provide further insight into the structure of the problem. Sections \ref{sec:upperbound} and \ref{sec:lowerbound} are devoted to the proof of the upper and lower bounds in our main theorem, respectively. We conclude in Section \ref{sec:conclusion}, where we also describe some interesting variations of our model, {which are further elaborated upon in the Appendix.}
\section{The Private Sequential Learning Model}
\label{sec:model}

We formally introduce our Private Sequential {Learning} model.  The model involves a \emph{{learner}} who aims to determine a particular \emph{true value}, $v^*$. The true value is a scalar in some bounded subset of $\mathbb{R}$. Without loss of generality, we assume that $v^*$ belongs to the interval\footnote{We consider a half-open interval here, which allows for a cleaner presentation, but the essence is not changed if the interval is closed.} $[0,  1)$ and that the {learner} knows that this is the case. The {true value} is stored in an external database. In order to learn the true value, the {learner} interacts with the database by submitting queries as follows. At each step $k$, the {learner} submits a \emph{query} $q_k\in[0,1)$, {and receives from the database a \emph{response}, $r_k$, indicating whether $v^*$ is greater than or equal to the query value}, i.e., 
\begin{equation*}
r_k = \mathbb{I}(v^*\geq q_k),
\end{equation*}
where $\mathbb{I}(\cdot)$ stands for the indicator function. 
Furthermore, each query is allowed to depend on the responses to previous queries, through a {learner} strategy, to be defined shortly. 

Denote by $N$ the total number of learner queries, and by $\epsilon>0$ the learner's desired accuracy. After receiving the responses to $N$ queries, the {learner} aims to produce an estimate ${\hat{x}}$, for $v^*$, that satisfies
\begin{equation*}
|{\hat{x}}-v^*|\leq\frac{\epsilon}{2}.
\end{equation*}  

In the meantime, there is an \emph{adversary} who is also interested in learning the {true value}, $v^*$. The adversary has no access to the database, and hence  seeks to estimate $v^*$ by free-riding on observations of the learner queries. Let $\delta>0$ be an accuracy parameter for the adversary. We assume that the adversary can observe the values of the queries but not the responses, and knows the {learner}'s query strategy.  
Based on this information, and after observing all of the queries submitted by the {learner}, the adversary aims to generate an estimate, $\hat{x}^a$, for $v^*$, that satisfies
\begin{equation*}
|\hat{x}^a-v^*|\leq\frac{\delta}{2}.
\end{equation*}

\subsection{{Learner} Strategy} 
\label{sec:user_strategy}

The queries that the {learner} submits to the database are generated by a (possibly randomized) \emph{{learner} strategy}, in a sequential manner: the query at step $k$ depends on the queries and their responses up until step $k-1$, as well as {on} a discrete random variable $Y$. In particular, the random variable $Y$ allows the {learner} to randomize if needed, and we will refer to $Y$ as the \emph{random seed}. Without loss of generality, we assume that $Y$ is uniformly distributed {over} $\{1,2,\dots,\mathcal{Y}\}$, where $\mathcal{Y}$ is a large integer. Formally, fixing $N\in\mathbb{N}$,
a {learner} strategy $\phi$ of length $N$ is comprised of two parts: 
\begin{enumerate}
 	\item A finite sequence of $N$ query functions, $(\phi_1, \dots, \phi_N)$, where each $\phi_k$ is a mapping that takes as input the values of the first $k-1$ queries submitted, the corresponding responses, as well as the realized value of $Y$, and outputs the $k$th query $q_k$.
 	\item An estimation function $\phi^{E}$, which takes as input the $N$ queries submitted, the corresponding responses, and the realized value of $Y$, and outputs the final estimate ${\hat{x}}$ for the true value $v^*$. 
 \end{enumerate} 
 More precisely, we have
\begin{enumerate}
 	\item If $k=1$, then $\phi_1\::\:{\{1,2,\dots,\mathcal{Y}\}}\rightarrow [0,1)$, and $q_1=\phi_1(Y)$;
 	\item[] If $k=2,3,\dots,N$, then $\phi_k\::\:[0,1)^{k-1}\times\{0,1\}^{k-1}\times {\{1,2,\dots,\mathcal{Y}\}}\rightarrow [0,1)$, and 
 	\begin{equation*}
 	q_k=\phi_k(q_1,q_2,\dots,q_{k-1},r_1,r_2,\dots,r_{k-1},Y);
 	\end{equation*}
 	\item 
 	$\phi^{E}\::\:[0,1)^{N}\times\{0,1\}^{N}\times {\{1,2,\dots,\mathcal{Y}\}}\rightarrow [0,1)$, and $\hat{x}=\phi^{E}(q_1,q_2,\dots,q_N,r_1,r_2,\dots,r_N,Y)$.
 \end{enumerate} 
Observe that the above definition can be simplified: knowing the value of the random seed $Y$ and the responses to the queries is sufficient for reconstructing the values of the queries. As an example, we have $q_2=\phi_2(q_1,r_1,Y)=\phi_2(\phi_1(Y),r_1,Y)=\phi_2'(r_1,Y)$, for some new function $\phi_2'$. Through induction, it then suffices to let the input to $\phi_k$ be just $r_1, \ldots, r_{k-1}$ and $Y$. This leads to an alternative, simpler definition of learner strategies:
 \begin{enumerate}
 	\item If $k=1$, then $\phi_1\::\:{\{1,2,\dots,\mathcal{Y}\}}\rightarrow [0,1)$, and $q_1=\phi_1(Y)$;
 	\item[] If $k=2,3,\dots,N$, then $\phi_k\::\{0,1\}^{k-1}\times {\{1,2,\dots,\mathcal{Y}\}}\rightarrow [0,1)$, and 
 	$q_k=\phi_k(r_1,r_2,\dots,r_{k-1},Y)$;
 	\item 
 	$\phi^{E}\::\{0,1\}^{N}\times {\{1,2,\dots,\mathcal{Y}\}}\rightarrow [0,1)$, and $\hat{x}=\phi^{E}(r_1,r_2,\dots,r_N,Y)$.
 \end{enumerate} 

{In what follows}, we adopt the latter, simpler definition. In addition, we will consider {learner} strategies that submit distinct queries, as repeated queries do not provide additional information to the {learner}.  We will denote by $\Phi_N$  the set of all {learner} strategies of length $N$, defined as above.

Fix a {learner} strategy $\phi \in \Phi_N$. To clarify the dependence on the random seed, {for any $x\in[0,1)$ and $y\in \{1,2,\dots,\mathcal{Y}\}$}, we will use $\overline{q}(x,y)$ to denote the realization of the sequence of  queries, {$(q_1,q_2, \ldots,q_N)$}, 
when the true value $v^*$ is $x$ and the {learner}'s random seed $Y$ is $y$. Similarly, we will denote by ${\hat{x}}(x,y)$ the {learner}'s estimate of the true value when $v^*=x$ and $Y=y$.  



\subsection{Information Available to the Adversary}
\label{sec:adv_info}
We summarize in this subsection the information available to the adversary. First, the adversary is aware that the true value $v^*$ belongs to $[0,1)$. Second, we assume that the adversary can observe the values of the queries but not the corresponding responses, and that the {learner} strategy $\phi$, {including the distribution of the random seed $Y$}, is known to the adversary. In particular, the adversary observes the value of each query $q_k${,} for $k=1,\dots,N$, and knows the $N$ mappings, $\phi_1,\phi_2,\dots,\phi_N$. This means that if the adversary had access to the values $r_1,r_2,\dots,r_{k-1}$ and the realized value of $Y$, she would know exactly what $q_k$ is for step $k$. {These assumptions stem from a worst-case consideration{:}  the privacy guarantees hold even when the adversary knows the learner's strategy, and can automatically extend to more practical scenarios where such knowledge may not be exact.} While it may seem that an adversary who sees both the {learner} strategy and her actions is too powerful to defend against, we will see {in the ensuing analysis} that the {learner} will still be able to implement effective and efficient obfuscation by exploiting the randomness {in} $Y$.  
\section{Private {Learner} Strategies}\label{sec:privatestrategycomplex}
In this section, we introduce and formally define private learning strategies, the central concept of this paper. While we will briefly discuss the underlying intuition, 
{further interpretation is provided in}
{Appendix \ref{sec:privacy_and_winning}}.  As was mentioned in {the Introduction}, a private {learner} strategy must always make sure that its estimate is close to the true value $v^*$, while keeping the {adversary}'s probability of 
{accurately estimating} $v^*$  sufficiently small. Our goal in this section is to formalize those ideas. To this end, we first introduce  in Section 3.1 {ways of quantifying} the amount of information acquired by  the adversary, as a function of the {learner}'s queries. This then leads to a precise privacy constraint{,} presented in Section 3.2.

\subsection{Information Set} \label{sec:infoconfiset}
Recall from Section \ref{sec:adv_info} that the adversary knows the values of the queries and the {learner} strategy. We will now convert this knowledge into a succinct representation: the \emph{information set} of the adversary. Fix a {learner} strategy, $\phi$. Denote by $\calQ(x)$ the set of query sequences that have a positive probability of appearing under $\phi$, when the true value $v^*$ is equal to $x$: 
\begin{equation}
\calQ(x)  =\{{\overline{q}} \in [0,1)^N: \pb_\phi({Q_x = {\overline{q}}}) >0 \},
\label{eq:calQx}
\end{equation}
{where {$Q_x$ is a vector-valued random variable representing the sequence of learner queries when the true value is equal to $x$, and {where} $\overline{q}$ 
{ranges over possible realized values;}
{the probability is measured with respect to the randomness in the learner's random seed, $Y$.}} 
\begin{definition} \label{def:infoset}
Fix $\phi\in\Phi_N$. 
The information set for the adversary, $\mathcal{I}({\overline{q}})$, is defined by: 
\begin{equation}
\mathcal{I}({\overline{q}}) = \Big\{x\in [0 , 1): {\overline{q}} \in \calQ(x) \Big\}, {\quad\overline{q}\in[0,1)^N.}
\end{equation}
\end{definition}
From the viewpoint of the adversary, the {information set} represents all possible {true values} {that are consistent with the queries observed}. As such, it captures the amount of information that the {learner} reveals to the adversary. 

\subsection{$(\epsilon,\delta,L)-$Private Strategies}\label{sec:stronger_privacy}
 
A private {learner} strategy should achieve two aims: accuracy and privacy. Accuracy can be captured in a relatively straightforward manner, by measuring the absolute distance between the {learner}'s estimate and the true value. An effective measure of the {learner}'s privacy, on the other hand, is more subtle, as it depends on what the adversary is able to infer. To this end, we develop in this subsection a privacy metric by quantifying the ``effective size'' of the information set $\calI({\overline{q}}) $ described in Definition \ref{def:infoset}. Intuitively, since the information set contains all possible realizations of the true value, $v^*$, the larger the information set, the more difficult it is for the adversary to pin down the true value. 

The choice of such a metric requires care. As a first attempt, the diameter of the information set, $\sup_{y_1,y_2\in \calI({\overline{q}})}{|y_1-y_2|}$, may appear to be a natural candidate. Since the adversary has an accuracy parameter of $\delta$, we 
{could impose, as a privacy constraint,} 
that the diameter of  $\calI({\overline{q}})$ be greater than $\delta$. The diameter, however, is not a good metric, as it paints an overly optimistic picture for the {learner}. 
{For example, consider the case} 
where the information set is the union of two intervals of length $\delta$ each, placed far apart from each other. By  setting her estimate to be the center of one of the two intervals, chosen at random with equal probabilities, the adversary will have probability $1/2$ of correctly predicting the true value, even though the diameter of the information set could be large.
{As a second attempt, we might consider} 
 the Lebesgue measure of the information set.  However, it also fails to {capture the intended meaning of} {learner} privacy. {For example, consider the case} where the information set consists of many distantly placed but very small intervals. It is not difficult to see that the adversary would not be able to correctly estimate the true value with high certainty, even if the Lebesgue measure of the set is arbitrarily small. 

The shortcomings {of} the above metrics motivate a more refined notion of ``effective size,'' and in particular, one that {would be appropriate for disconnected information sets}. To this end, we will use set coverability to measure the size of the information set, defined as follows.

\begin{definition}
\label{def:coverable}
Fix $\delta>0$, $L\in\mathbb{N}$, and a set $\calE\subset \R$. We say that a collection of $L$ closed intervals $[a_1,b_1]$, $[a_2,b_2]$, $\dots$, $[a_L,b_L]$, is a $(\delta,L)$ cover for $\calE$ if  $\mathcal{E}\subset\bigcup_{1\leq j\leq L}[a_j,b_j]$,  and $b_j - a_j \leq \delta$ for all $j.$ 

We say that a set $\calE$ is {\bf $(\delta,L)$-coverable} if it admits {a} $(\delta,L)$ cover. {In addition, we define the {\bf $\delta$-cover number} of a set $\mathcal{E}$, $C_\delta(\mathcal{E})$, as
\begin{equation}
C_\delta(\mathcal{E})\triangleq\min\: \{L\in\mathbb{N}: \textrm{ $\mathcal{E}$ is $(\delta,L)$-coverable}\}.
\end{equation}}
\end{definition}

We are now ready to define $(\epsilon,\delta,L)$-private {learner} strategies.
\begin{definition}[Private {Learner} Strategy]
\label{def:private_strategy}
Fix $\epsilon>0$, $\delta>0$, $L\geq 2$, with $L\in\mathbb{N}$. A {learner} strategy $\phi\in\Phi_N$ is $(\epsilon,\delta, L)$-private if it satisfies the following: 
\begin{enumerate}
\item Accuracy constraint: the {learner} estimate accurately recovers the true value, with probability one:
\begin{equation*}
\mathbb{P}\Big( \big|\hat{x}(x,Y)-x\big| \leq {\epsilon}/{2}\Big)=1, \quad \forall\: x\in[0,1), 
\end{equation*}
where the probability is measured with respect to the randomness in $Y$. 
	\item Privacy constraint: for every $x\in[0,1)$ and every possible sequence of queries $\overline{q} \in \calQ(x)$, {the $\delta$-cover number of the {information set} for the adversary, $C_\delta\big(\calI(\overline{q})\big)$, is 
	at least $L$, i.e., 
	\begin{equation}
	C_\delta\big(\calI(\overline{q})\big)\geq L,\quad\forall\: \overline{q}\in\mathcal{Q}(x).
	\end{equation} } 
\end{enumerate}
\end{definition}

The accuracy constraint requires that a private {learner} strategy always produce an accurate estimate within the error tolerance $\epsilon$, for any possible true value in $[0,1)$. The privacy constraint controls the size of the information set induced by the sequence of queries generated, and the parameter $L$ can be interpreted as the {learner}'s privacy level: since the intervals used to cover the information set are of length at most $\delta$, each interval can be thought of as representing a plausible guess for the adversary. Therefore, the probability of the adversary successfully estimating the location of $v^*$ is essentially inversely proportional to the number of intervals needed to cover the information set, which is at most $1/L$.  {It turns out that this intuition can be made precise: {in Appendix \ref{sec:privacy_and_winning},} we formally establish the equivalence between $1/L$ and the adversary's probability of correct estimation}.

\subsection{{Worst-Case versus Bayesian Formulations}}
\label{sec:risk_averse_adversary}

{Our definition of learner privacy involves worst-case requirements for both the learner and the adversary. In a Bayesian formulation, these are replaced by requirements that only need to hold on the average, under a prior distribution for the value of the unknown target $v^*$. We formulate such a Bayesian variant in detail in Appendix \ref{app:baysianprivate}, and argue that it is complementary, not directly comparable, to our main formulation.  On the technical side, in Appendix \ref{app:baysianprivate} we present a learner strategy that achieves privacy with a query complexity that depends \emph{multiplicatively} on $L$; in more recent work that follows up on the current paper, \cite{kuangNIPS2018} establishes a lower bound that shows that such dependence is also tight. We note that not only the two formulations are not comparable, but their analysis is also different: 
the lower bounds for our base model rely  on combinatorial arguments, in contrast to information-theoretic arguments for the Bayesian variant \citep{kuangNIPS2018}.}


\section{Main Result}\label{sec:complex_results}
The {learner}'s overall objective is to employ {a minimal} number of queries{,} while satisfying the accuracy and privacy requirements. We state our main theorem in this section, which establishes lower and upper bounds for the query complexity of a private {learner} strategy, as a function of the adversary accuracy $\delta$, {learner} accuracy $\epsilon$, and {learner} privacy level, $L$. Recall that $\Phi_N$ is the set of {learner} strategies of length $N$. {We define} $N^*(\epsilon,\delta,L)$ {as} the minimum number of queries needed across all $(\epsilon,\delta,L)$-private {learner} strategies, 
\begin{equation}
\label{eqn:definition of N2}
N^*(\epsilon,\delta,L)=\min\big\{N\in\mathbb{N}: \text{$\Phi_N$ contains at least one $(\epsilon,\delta,L)$-private strategy}\big\}. 
\end{equation}

Our result {focuses} on the regime of parameters where 
\begin{equation}
0<2\epsilon<\delta\leq1/L. 
\end{equation}
\begin{enumerate}
\item[(i)]
Having $2\epsilon<\delta $ corresponds to a scenario where the {learner} {wants to estimate} the true value with high accuracy, while the adversary is {content with} a coarse estimate.  {In contrast,} the regime where $\delta<\epsilon$ is arguably much less interesting{:}  it is not natural {for} the adversary, who is not engaged in the querying process, to {aim at} higher accuracy than the {learner}. 
\item[(ii)]
The requirement that $\delta\leq1/L$ stems 
{from the 
following argument. If $\delta\geq 1/(L-1)$, then 
the entire interval $[0,1)$ is trivially $(\delta,L-1)$-coverable, and  $C_\delta\big(\calI(\overline{q})\big)
\leq C_\delta\big([0,1)\big)\leq L-1 <L$. Thus, the privacy constraint is automatically  violated, and no private {learner} strategy exists. 
To obtain a nontrivial problem, we therefore only need to consider the case where $\delta< 1/(L-1)$, which is only sightly broader than the regime $\delta\leq 1/L$ that we consider. 
}
\end{enumerate}
The following theorem is the main result of this paper.

\begin{theorem}[Query Complexity of Private Sequential Learning]
\label{thm:1}
Fix $\epsilon>0$, $\delta>0$, and a positive integer $L\geq 2$, such that $2\epsilon< \delta\leq{1}/{L}$. Then,
\begin{equation} 
\max\Big\{\log\frac{1}{\epsilon},\:\log\frac{\delta}{\epsilon}+2L-4\Big\} \leq N^*(\epsilon,\delta,L)\leq \log\frac{1}{L\epsilon}+2L.
\end{equation}
\end{theorem} 

The proof of  the upper bound in Theorem \ref{thm:1} is constructive, providing a specific {learner} strategy that satisfies the bound. 
If we set $\delta={1}/{L}$,  where the adversary's accuracy requirement is essentially as loose as possible, {and thus  corresponds to a worst case for the learner,} then Theorem \ref{thm:1} leads to the following  corollary, {which} yields upper and lower bounds that are tight up to an additive constant of~$4$. In other words, the private {learner} strategy that we construct  achieves essentially the optimal query-complexity in this scenario. 

\begin{corollary} \label{cor:1}
Fix $\epsilon>0$ and a positive integer $L\geq 2$ such that $2\epsilon< 1/L$. The following holds. 
\begin{enumerate}
	\item If $L=2$, {then}
	\begin{equation}
\begin{split}
\log\frac{1}{\epsilon} \leq N^*\Big(\epsilon,\frac{1}{L},L\Big)\leq \log\frac{1}{\epsilon}+4.
\end{split}
\end{equation}
\item If $L\geq 3$, {then}
\begin{equation} \label{theorem_bound_L}
\begin{split}
\log\frac{1}{L\epsilon}+2L-4 \leq N^*\Big(\epsilon,\frac{1}{L},L\Big)\leq \log\frac{1}{L\epsilon}+2L.
\end{split}
\end{equation}
\end{enumerate}
\end{corollary} 

A  {key message} from the above results is about the price of privacy: it is not difficult to see that in the absence of a privacy constraint, the most efficient {strategy, using a bisection search,} can locate the true value with $\log({1}/{\epsilon})$ queries. Our results thus demonstrate that the {price of privacy is at most an \emph{additive} factor of $2L$.}

{We close by noting the following two important aspects of our upper bounds:}
\begin{enumerate}
	\item Randomization: the proof of our upper bounds involves a strategy that relies strongly on the availability of the randomization variable $Y$. It is not known whether a  deterministic private learner strategy, with comparable query complexity, is possible.
	\item {Practical relevance  and limitation: while mathematically valid, the particular strategy we developed heavily exploits the underlying structure of the information set. When the learner's desired accuracy, $\epsilon$, is very small, the resulting information set will contain separate, tiny ``guesses'' (intervals). From an applied perspective, the chance that these tiny guesses contain $v^*$ is negligible, and a practical adversary might benefit by essentially ignoring them, instead of having to ``cover'' them under the current worst-case formulation. We postpone a detailed discussion together with two potential remedies until the Conclusion (Section \ref{sec:conclusion}), after developing a necessary understanding of the proof in the next sections. }
\end{enumerate}

\section{Examples of {Learner} Strategies}

\label{sec:example_strategy}

Before delving into the proofs of our main result, we first provide some intuition and motivation by examining three representative {learner} strategies situated at different locations along the complexity-privacy tradeoff curve.  


{\bf Strategy 1: Bisection.}  A most natural candidate is the classical bisection strategy, which is known to achieve the optimal query-complexity in the absence of privacy constraints. Under this strategy, the {learner} first submits a query at the mid-point of $[0, 1)$, i.e., $q_1=0.5$. Then, based on the response, the {learner} identifies the half interval that contains the true value, and subsequently submits its mid-point as the next query, $q_2$. The process continues recursively until the {learner} finds an interval of length at most $\epsilon$ that contains the true value $v^*$. Figure \ref{strategy1} provides an illustration of this strategy. 
\begin{figure}[h]
\vspace{-0.15in}
 \centering
   \includegraphics[width=6.2in]{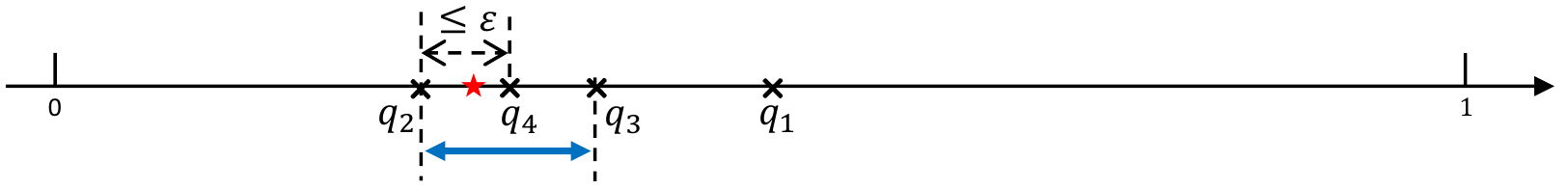}
   \caption{An example of the {B}isection {strategy.} {The} red star represents the true value $v^*$. The dashed line with arrows represents the {learner}'s error tolerance. {The} solid line with arrows represents the information set {of} the adversary, $\calI({\overline{q}})$. }\label{strategy1}
   \vspace{-0.25in}
 \end{figure}

Under the Bisection strategy, the {learner} knows that the interval containing the true value is halved with each successive query. It follows that the number of queries needed under the bisection strategy is $N=\log(1/\epsilon)$.  Unfortunately, the favorable query complexity afforded by the bisection strategy comes at the cost of the {learner}'s privacy. In particular, at the end of the process, the adversary knows that the true value must be close {(within $\epsilon$)} to the  last query the {learner} submitted.
{In particular, the information set is an interval of length at most $2\epsilon$.
Hence, under our assumption that $\delta>2\epsilon$, its
$\delta$-cover number is 1, and the strategy is not private, for any $L\geq 2$.} The  Bisection strategy lies at one extreme end of the complexity-privacy tradeoff, with a minimal query complexity but no privacy. 

{\bf Strategy 2: $\epsilon$-Dense.} {At} the opposite end of the spectrum is the $\epsilon$-Dense strategy, where the {learner} submits a {pre-determined} sequence of $N=1/\epsilon-1$  queries, with $q_1 = \epsilon$, $q_2 = 2\epsilon, \ldots, q_N = N\epsilon$ (Figure \ref{strategy2}). The strategy is accurate because the distance between {any} two adjacent queries is equal to the error tolerance, $\epsilon$. Moreover, because the sequence of queries is pre-determined and does not depend on the location of the true value, the adversary obtains no information from the {learner}'s query {pattern,}  and the information set remains the interval $[0,1)$ throughout. Thus, {as long as} $\delta\leq 1/L$, the strategy is $(\epsilon, \delta, L)$-private. Compared to the Bisection strategy, the perfect privacy of $\epsilon$-Dense Strategy is achieved at the expense of an \emph{exponential} increase in query complexity,  from $\log(1/\epsilon) $  to $1/\epsilon$. The $\epsilon$-Dense strategy is therefore overly conservative and, as our proposed strategy will demonstrate, leads to unnecessarily high query complexity for moderate values of $L$.

\begin{figure}[h]
 \centering
   \includegraphics[width=6in]{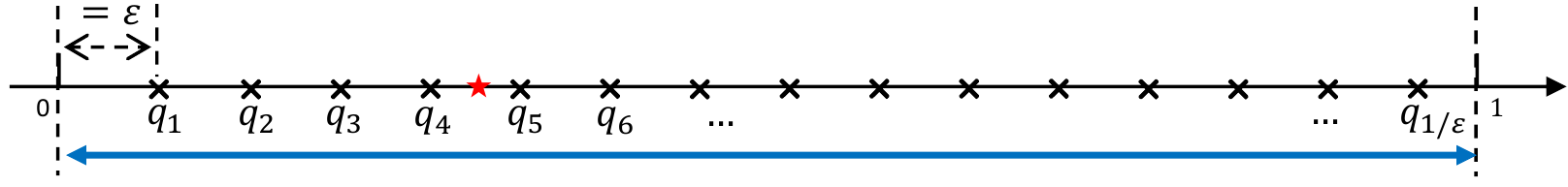}
   \caption{An example of the $\epsilon$-Dense strategy. The dashed line with arrows represents the {learner}'s error tolerance. {The} solid line with arrows represents the information set {of} the adversary, $\calI({\overline{q}})$. }\label{strategy2}
   \vspace{-0.25in}
 \end{figure}

{\bf Strategy 3: Replicated Bisection.}
The contrast between Strategies 1 and 2  highlights the tension between the learner's conflicting objectives: on the one hand, to  maximally exploit the information learned from earlier queries and shorten the {search}, and on the other hand, to reduce adaptivity so that the queries are not too revealing. An efficient private {learner} strategy should therefore strike a balance between these two objectives.  To start, it is natural to consider a {learner} strategy that combines Strategies 1 and 2 in an appropriate manner, which leads us to the Replicated Bisection strategy, {which we describe next, and which consists of} two phases:
\begin{enumerate}
\item \emph{Phase 1 - Deterministic Queries}. The {learner} submits $L-1$ queries, chosen deterministically: 
 \begin{equation}
 q_1 = \frac{1}{L},\ q_2 = \frac{2}{L}, \ldots,\: q_{L-1} = \frac{L-1}{L}. 
 \end{equation}
{These queries} partition the unit interval into $L$ disjoint sub-intervals of length $1/L$ each, {namely,} $[0,1/L), [1/L, 2/L), \ldots, [1-1/L, 1)$.  At this point, the {learner} {can determine}  which one of the $L$ sub-intervals contains the true value, while the adversary has gained no additional information about the true value. We will refer to the sub-interval that contains the true value as the \emph{true sub-interval}, and all other sub-intervals as \emph{false sub-intervals}. This phase uses $L-1$ queries. 
\item \emph{Phase 2 - Replicated Bisection}. In the second phase, the {learner} conducts a bisection strategy within the true sub-interval until the true value has been located, while in the meantime {submitting translated replicas of these queries in each} false sub-interval, in parallel. The exact order in which these queries are submitted can be arranged in such a manner as to be independent from the identity of the true sub-interval. This phase uses $L\log(1/L\epsilon)$ queries, where $\log(1/L\epsilon)$ is the number of queries needed to conduct a bisection strategy in a sub-interval. 
\end{enumerate} 

When the process is completed, the {learner} will have identified the true value via the bisection strategy within the true sub-interval, while the adversary will have seen $L$ identical copies of the same bisection {strategy}, leading to an information set that consists of $L$ disjoint length-$2\epsilon$  intervals,  separated from each other by a distance of $1/L - 2\epsilon$.
It is not difficult to show that the Replicated Bisection strategy is $(\epsilon, \delta, L)$-private, with $L\log(1/L\epsilon) + L-1$ queries. In particular, the {Replicated Bisection strategy} achieves privacy at the cost of an increase in query complexity that is a \emph{multiplicative} factor of $L$, compared to that of the Bisection strategy ($N=\log(1/\epsilon)$).

\begin{figure}[h]
 \centering
   \includegraphics[width=6in]{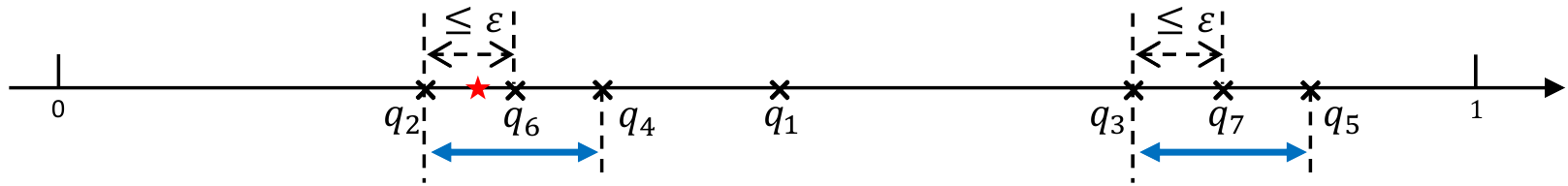}
   \caption{An example of the Replicated Bisection strategy, with $L=2$. The dashed lines with arrows {represent} the {learner}'s error tolerance.  {The} solid lines with arrows {represent} the information set {of} the adversary, $\calI({\overline{q}})$. }\label{strategy4}
   \vspace{-0.25in}
 \end{figure}

The Replicated Bisection strategy thus appears to be a natural and successful combination of the Bisection and $\epsilon$-Dense strategies: it ensures privacy while requiring substantially fewer than the $N=1/\epsilon$ queries of the $\epsilon$-Dense strategy. {Nevertheless, the upper bound in Theorem \ref{thm:1} indicates that the query complexity  can be significantly improved, with an additive --- rather than multiplicative --- dependence on $L$.}

\section{Proof of the Upper Bound: Opportunistic Bisection Strategy}
\label{sec:upperbound}

We prove in this section the upper bound on the query complexity in Theorem \ref{thm:1}. This is achieved by constructing a specific {learner} strategy, which we will refer to as the 
Opportunistic Bisection (OB) strategy. We start with some terminology, to facilitate the exposition. 

\begin{definition} Fix $M \in \N$ and an interval $\calJ \subset [0,1)$. Let $Z = (Z_1, Z_2, \ldots)$ be an infinite sequence of i.i.d.~Bernoulli random variables, with $\pb(Z_i=0)=1/2$. Let $(q_1,q_2, \ldots, q_M)$ be a sequence of $M$ queries, where $q_1$ is equal to the mid-point of $\calJ$, and let $(r_1, r_2, \ldots,r_M)$ be their corresponding responses. 
\begin{enumerate}
\item We say that $(q_1,q_2, \ldots, q_M)$ is a {\bf truthful bisection search of $\calJ$}, if it satisfies the following criteria, defined inductively. Let $\calJ_1 = \calJ$. For $i = 1,2, \ldots, M$,
\begin{enumerate}[(a)]
\item $q_i$ is set to 
\begin{equation}
 q_i = \mbox{mid-point of interval $ \calJ_i$}.
 \end{equation} 
 \item $\calJ_{i+1}$ is set to 
\begin{equation}
\calJ_{i+1}= \left\{
\begin{array}{ll}
\left[\inf \calJ_i  \, ,  \, q_i \right),  & \quad \mbox{if $r_i=0$}, \\
\left[q_i  \, ,  \,  \sup \calJ_i \right), & \quad \mbox{if $r_i=1$}.
\end{array}
\right.
\end{equation}
\end{enumerate}
\item We say that $(q_1,q_2, \ldots, q_M)$ is a {\bf fictitious bisection search of $\calJ$}, if it satisfies the following criteria, defined inductively. Let $\calJ_1 = \calJ$. For $i = 1,2, \ldots, M$,
\begin{enumerate}[(a)]
\item  $q_i$ is set to
\begin{equation}
 q_i = \mbox{mid-point of interval $ \calJ_i$}.
 \end{equation} 
 \item $\calJ_{i+1}$ is set to 
\begin{equation}
\calJ_{i+1}= \left\{
\begin{array}{ll}
\left[\inf \calJ_i  \, ,  \, q_i \right),  & \quad \mbox{if $Z_i=0$}, \\
\left[q_i  \, ,  \,  \sup \calJ_i \right), & \quad \mbox{if $Z_i=1$}.
\end{array}
\right.
\end{equation}
\end{enumerate}
\end{enumerate}
\end{definition}
In words, whether a bisection search is truthful or fictitious depends on how the interval $\calJ_i$ is updated. In a truthful search, $\calJ_{i+1}$ is set to the half-interval within $\calJ_i$ that, according to the response $r_i$, contains the true value. In a fictitious search, this choice is made uniformly at random, according to $Z$. 

We are now ready to define  the Opportunistic Bisection strategy, which consists of two phases.

\emph{Phase 1 - Opportunistic Guesses}. The first $2L$ queries submitted by the strategy are deterministic and do not depend on responses from earlier queries, with
\begin{equation}
q_i = (i-1)\frac{1}{L}, \quad i = 1, \ldots, L,
\end{equation}
and 
\begin{equation}
(q_{L+1}, q_{L+2}, \ldots, q_{2L}) = (q_1+\epsilon, q_2+\epsilon,\ldots, q_L+\epsilon).
\end{equation}

Notice that the two queries $q_i$ and $q_{i+L}$ determine an interval $[q_i,q_{i+L})$ of length $\epsilon$. At the end of this phase, there will be  $L$ such intervals, evenly spaced across the unit interval. Each such interval $[q_i,q_{i+L})$ thus represents a ``guess'' on the true value, $v^*$; if $v^*$ lies in $[q_i,q_{i+L})$ for some $i\in \{1,\ldots, L\}$, then the {learner} learns the location of $v^*$ within the desired level of accuracy. We will refer to the interval $[q_i,q_{i+L})$ as the \emph{$i$th guess}.

\emph{Phase 2 - Local Bisection Search}. The guesses submitted in Phase 1 are few and spaced apart, and it is possible that none of the $L$ guesses contains $v^*$. The goal of Phase 2 is to hence ensure that the {learner} identifies $v^*$ at the end, but the queries are to be executed in a fashion that conceals from the adversary whether $v^*$ was identified during Phase 1 or Phase 2. 

Define $\calJ^{(i)}$ as the interval between the $i$th and $(i+1)$th guesses: 
\begin{equation}
\calJ^{(i)} = [q_{L+i}, \, q_{i+1})  = \lt[(i-1)\frac{1}{L}+\epsilon, \, \frac{i}{L} \rt), \quad i = 1, 2, \ldots, L. 
\end{equation}
We will refer to $\calJ^{(i)}$ as the \emph{$i$th sub-interval}. Importantly, by the end of Phase $1$, if none of the guesses contains the true value, then the {learner} knows which sub-interval contains the true value, which we will denote by $\calJ^*$. The queries in Phase 2 will be chosen according to the following rule: 
\begin{enumerate}
\item If none of the guesses in Phase $1$ contains $v^*$, then, let $(q_{2L+1}, q_{2L+2}, \ldots, q_{2L+M})$  be  a {\bf truthful bisection  search} of $\calJ^*$ with $M=\log\lt( \frac{1}{\epsilon L} \rt)$. 

\item If one of the guesses in Phase $1$ contains $v^*$, then, let $\tilde{\calJ}$ be a sub-interval chosen uniformly at random among all $L$ sub-intervals, and let $(q_{2L+1}, q_{2L+2}, \ldots, q_{2L+M})$  be  a {\bf fictitious bisection search} of $\tilde\calJ$ with $M=\log\lt( \frac{1}{\epsilon L} \rt)$, using the randomization provided by $Y$ (i.e., using $Y$ to generate the sequence of i.i.d. Bernoulli random variables, $Z_i$). 
\end{enumerate}
An example of this strategy is provided in Figure \ref{fig:OB}.

\emph{Remark.} It is interesting to contrast Opportunistic Bisection with the Replicated Bisection strategy in Section \ref{sec:example_strategy}. Both strategies use deterministic queries in the first phase, but instead of submitting $L$ queries, the OB strategy incurs a slight overhead and submits $L$ \emph{guesses} {($2L$ queries)}. Crucially, the guesses make it possible to immediately discover the location of the true value in the first phase, albeit such discoveries might be unlikely. In the second stage, while the Replicated Bisection strategy conducts a bisection search in \emph{each} of the $L$ sub-intervals, the Opportunistic Bisection strategy does so in only \emph{one} of the sub-intervals, hence drastically reducing the number of queries. 

It follows directly from the definition that the number of queries submitted under the Opportunistic Bisection strategy is: 
\begin{equation}
N = 2L + \log\lt(\frac{1}{\epsilon L} \rt). 
\end{equation}

\begin{figure}[htp]
 \centering
   \includegraphics[scale=0.78]{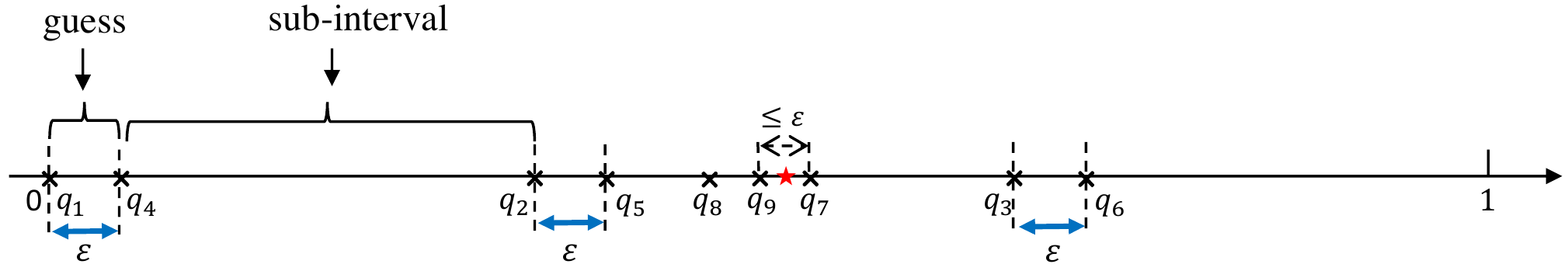}
   \caption{An example of the Opportunistic Bisection Strategy, with $L=3$.}
   \label{fig:OB}
 \end{figure}

To complete the proof of the upper bound in Theorem \ref{thm:1}, it thus suffices to show that the OB strategy satisfies both the accuracy and privacy constraints. This is accomplished in the following proposition, which is the main result of this section. 
\begin{proposition}
\label{prop:upperbound}
Fix $\epsilon>0$, $\delta>0$, and a positive integer $L\geq 2$, such that $2\epsilon< \delta\leq{1}/{L}$. Then, the Opportunistic Bisection (OB) strategy is  $(\epsilon, \delta, L)$-private. 
\end{proposition}

\bpf We first show that the OB strategy is accurate, and specifically, that it will allow the {learner} to produce an estimate of $v^*$ with an absolute error of at most $\epsilon/2$. To this end, we consider two possible scenarios:  

\emph{Case 1}. Suppose that some guess in Phase $1$, namely, the interval $[q_{i'}, q_{i'+L})$,  contains the true value, $v^*$. In this case, the {learner} can set $\hat{x}$ to be the mid-point of the guess, i.e., $\hat{x} = (q_{i'} + q_{i'+L})/2$. Since the length of each guess is exactly $\epsilon$, we {will} have $|\hat{x}-v^*|\leq \epsilon/2$. 

\emph{Case 2}. Suppose that none of the guesses in Phase $1$ contains $v^*$. This means that a \emph{truthful} bisection search will be conducted in Phase 2, in the sub-interval that contains $v^*$. Because the search is truthful, we know that one of the two intervals adjacent to $q_N$ must contain $v^*$. Let this interval be denoted by $H^*$. Furthermore, because the length of each sub-interval is less than $1/L$ and there are $\log(1/\epsilon L)$ steps in the bisection search, we know that the length of $H^*$ is at most $\epsilon$. Therefore, the {learner} can generate an accurate estimate by setting  $\hat{x}$ to be the mid-point of $H^*$. Together with Case 1, this shows that the OB strategy leads to an accurate estimate of $v^*$. 

We now show that the OB strategy is private, and in particular, that {the $\delta$-cover number of the information set of the adversary, $C_\delta\big(\calI(\overline{q})\big)$,  is at least $L$}. Denote by $\calG$ the union of the guesses, i.e., 
\begin{equation}
\calG = \bigcup_{i=1}^L \, [q_i,q_{i+L}).
\end{equation}

It is elementary to show that for two sets $U$ and $V$, with $U \subset V$, {if $C_\delta(U)$ is at least $L$, then so is $C_\delta(V)$}. Therefore, it suffices to prove the following two claims. 
\begin{claim} 
\label{clm:1}
{The $\delta$-cover number of $\calG$, $C_\delta(\calG)$, is at least $L$.}
\end{claim}

\begin{claim} 
\label{clm:2}
The information set, $\calI(\overline{q})$, contains $\calG$. 
\end{claim}

We first show Claim \ref{clm:1}. Consider any interval $J\subset [0,1)$, with length {at most} $\delta$, used in a cover for $\calG$. Note that by construction, each guess has length $\epsilon$, and two adjacent guesses are separated by a distance of $1/L-\epsilon$. Since $\delta\leq 1/L$, this implies that the Lebesgue measure of $J\cap \calG$ is at most $\epsilon$. Since the Lebesgue measure of $\calG$ is $\epsilon L$, we conclude that it will {take} at least $L$ such intervals $J$ {(i.e.,} of size {at most} $\delta${)} to cover $\calG$. Therefore, {$C_\delta(\calG)\geq L$}. This proves Claim \ref{clm:1}.

We next show Claim \ref{clm:2}.
Note that  any particular query sequence $\overline{q}$ can arise in two different ways: (i) it may be that $v^*$ is an arbitrary element of one of the guesses (i.e., $v^*\in \cal{G}$), and  $\overline{q}$ is the result of a fictitious bisection search; or, (ii) it may be that $v^*$ lies outside the guesses, and $\overline{q}$ is the result of a truthful bisection search.
The adversary has no way of distinguishing between these two possibilities. Furthermore, there is no information available to the adversary that could distinguish between different elements of $\cal{G}$. As a consequence, all elements of $\cal{G}$ are included within the information set, i.e., 
${\cal G}\subset {\cal I}(\overline{q})$, which establishes  Claim \ref{clm:2}. \qed


\section{Proof of the Lower Bound}\label{sec:lowerbound}

We now derive the two lower bounds {in Theorem \ref{thm:1}, on the} query complexity. {Note that the query-complexity of the Opportunistic Bisection strategy carries a $2L$ overhead compared to the (non-private) bisection strategy. The value $2L$ admits an intuitive justification: a private learner strategy must create $L$ plausible locations of the true values, and each such location is associated with at least 2 queries. One may question, however, whether the $2L$ queries need to be \emph{distinct} from the $\log(1/\epsilon)$ queries already used by the bisection search, {or whether} the query complexity could be further reduced by ``blending'' the queries for obfuscation with those for identifying the true value in a more effective manner. The key to the proof of the lower bound in this section is to show that such ``blending'' is not possible: 
{in order to successfully obfuscate the true value, one needs $2L$ queries that are \emph{distinct} from those that participate in the bisection algorithm.}
}

\subsection{Information Sets}
We first introduce some notation to facilitate our discussion. Recall that $(q_1,q_2, \ldots, q_N)$ is the sequence of {learner} queries. For the remainder of this section, we augment this sequence with two more queries, $q_0 {\:\triangleq\:} 0$ and $q_{N+1} {\:\triangleq\:} 1$, so that ${\overline{q}} = (0, q_1,q_2, \ldots, q_N, 1)$. This is inconsequential because  $0\leq v^*< 1$, and hence adding $q_0$ and $q_{N+1}$ does not provide additional information to either the {learner} or the adversary. 

We start by examining the information provided to the learner, through the queries and the responses. 
Let us fix an arbitrary $y\in \{1,\ldots,\cal Y\}$ that has positive probability, and some $v^*\in[0,1)$.
Consider the resulting sequence of queries, $\overline{q}=\overline{q}(v^*,y)$, and then let
$\overline{q}^S=(q^0,q^1,\ldots,q^N, q^{N+1})$ be the sequence of queries in $\overline{q}$, arranged in increasing order. (In particular, $q^0=0$ and $q^{N+1}=1$.) For each query $q^i$, the learner knows (through the response to the corresponding query) whether $v^*<q^i$ or $v^*\geq q^i$. In particular, at the end of the learning process, the learner has access to an interval of the form $H=[q^i,q^{i+1})$, for some $i\in \{0,1,\ldots, N\}$, such that $v^*$ is certain to belong to that interval. Furthermore, from the definition of learner strategies, all elements of that interval would have produced identical responses to the queries, and the learner has no information that distinguishes between such elements. 

It is not hard to see that if $q^{i+1}-q^i>\epsilon$, then the learner has no way of producing an $(\epsilon/2)$-accurate estimate of $v^*$.\footnote{For any choice of $\hat x$, there will always be some $x\in H$ such that $|\hat x-x|>\epsilon/2$. Furthermore, such an $x$ is a possible value of $v^*$, as it would have produced the exact same sequence of responses.} Since we are interested in learner strategies that satisfy the accuracy constraint in Definition \ref{def:private_strategy}, we conclude that the length of $H$ is at most $\epsilon$.

Let us now consider the situation from the point of view of the adversary. The adversary can look at the query sequence $\overline{q}$, form the intervals of the form $[q^i,q^{i+1})$, and select those intervals whose length is at most $\epsilon$; we 
refer to these as \emph{special intervals}. We have already argued that $v^*$ must lie inside a special interval. Therefore, the adversary has enough information to conclude  that $v^*$ lies in the union of the special intervals. We denote that union by
$\overline{{\cal I}}(\overline{q})$, and we have
\begin{equation}\label{eq:ii}
{\cal I}(\overline{q}) \subset \overline{{\cal I}}(\overline{q}).
\end{equation}


\subsection{Completing the Proof of the Lower Bound}
\label{sec:completeLowerBound}
We are now ready to prove the lower bound. We begin with a lemma.
\begin{lemma} 
\label{lem:genBiComp}
Fix a {learner} strategy $\phi$ that satisfies the accuracy constraint. 
For every $y\in \{1,2,\ldots, \calY\}$,  there exists $x \in (0,\delta)$ such that
there are at least $\log(\delta/\epsilon)$ of the queries in $\overline{q}(x,y)$ that belong to $(0,\delta)$.
\end{lemma}
Lemma \ref{lem:genBiComp} is essentially the classical result that $\log(1/\epsilon)$ query complexity {of the bisection strategy is optimal} for the unit interval (cf.~\cite{waeber2013bisection}), which proves the first term of the lower bound in Theorem \ref{thm:1}. We omit the proof of Lemma \ref{lem:genBiComp}, which is fairly standard, but provide an intuitive argument. Fix $y\in \{1,2,\ldots, \calY\}$. Note that the interval $(0,\delta)$ consists of $\delta/\epsilon$ disjoint sub-intervals of length $\epsilon$ each. An accurate {learner} strategy, therefore, must be able to distinguish in which one of these sub-intervals the true value resides. Distinguishing among $\delta/\epsilon$ possibilities using binary feedback therefore implies that there will be some $v^* \in (0,\delta)$ whose accurate identification requires $\log(\delta/\epsilon)$ queries in $(0,\delta)$. 

For the rest of this proof, fix $v^*=x_0$ and $Y=y_0$ for some $x_0$ and $y_0$ satisfying Lemma \ref{lem:genBiComp}, and use $\overline{q}$ to denote $\overline{q}(x_0,y_0)$. We now consider the queries in the interval $[\delta,1]$. Among them, we restrict attention to those queries that are endpoints of special intervals. We call these \emph{special} queries, and let $K$ be their number. We sort the special queries in ascending order, and denote them by 
 $s_1,s_2,\ldots,s_K$, where $K$ is their number. 
 
The outline of the rest of the argument is as follows. For a private learner strategy, the $\delta$-cover number of ${\cal I}(\overline{q})$ is at least $L$. From Eq. (\ref{eq:ii}), it follows that the $\delta$-cover number of $\overline{\cal I}(\overline{q})$ is also at least $L$. 
Since endpoints of special intervals are within $\epsilon$ of each other, 
every $s_i$ must be within $\epsilon$ of a ``neighboring'' query (namely, $s_{i-1}$ or $s_{i+2}$), with the possible exception of $s_1$, which could be the right endpoint of a special interval whose left endpoint, denoted $s_0$, is in $[0,\delta)$. Using the assumption that $\delta>2\epsilon$, each interval used in the cover can include two, and often three, queries $s_i$. In what follows, we will make this argument precise, and show that the number, $K$, of special queries is  at least $2L-3$.

Let us consider first the case where the above mentioned exception does not arise; i.e., we assume that $s_1$ is \emph{not} the right endpoint of a special interval. 
We decompose the set $\overline{\cal I}(\overline{q})\cap [\delta,1]$ as the union of its (finitely many) connected components, which we will call just \emph{components}, for short; see Figure \ref{lowerboundaa} for an illustration.
Each connected component
is an interval whose endpoints are special queries ($s_i$, for some $i$). Furthermore within each such interval, special queries are separated by at most $\epsilon$.
Suppose that we have $m$ components. For the $j$th component, let $k_j$ be the number of special queries it contains, and let $d_j$ be its length.
We have $d_j\leq (k_j-1)\epsilon$. In particular, the $j$th interval can be covered by at most
$$\Big\lceil \frac{d_j}{\delta}\Big\rceil \leq 
\Big\lceil (k_j-1) \frac{\epsilon}{\delta}\Big\rceil 
\leq \Big\lceil \frac{k_j-1}{2} \Big\rceil \leq \frac{k_j}{2}$$
intervals of length $\delta$. (We have used here our standing assumption that $\delta>2\epsilon$.
Summing over the different components, we conclude that 
$\overline{\cal I}(\overline{q})\cap [\delta,1]$
can be covered by at most
$\sum_j k_j/2=K/2$ intervals of length $\delta$. 

For the exceptional case where $s_1$ is the right endpoint of a special interval $[s_0,s_1]$, we just apply the same argument, now on the set $\overline{\cal I}(\overline{q})\cap [s_0,1]$, and for an augmented collection of special queries,  $(s_0,s_1,\ldots,s_K)$. 
Having effectively increased the number of points of interest, from $K$ to $K+1$, we obtain an upper bound of $(K+1)/2$ on the number of intervals of length $\delta$ that are needed to cover $\overline{\cal I}(\overline{q})\cap [s_0,1]$.

 \begin{figure}[H]
 \centering
   \includegraphics[scale=0.8]
   {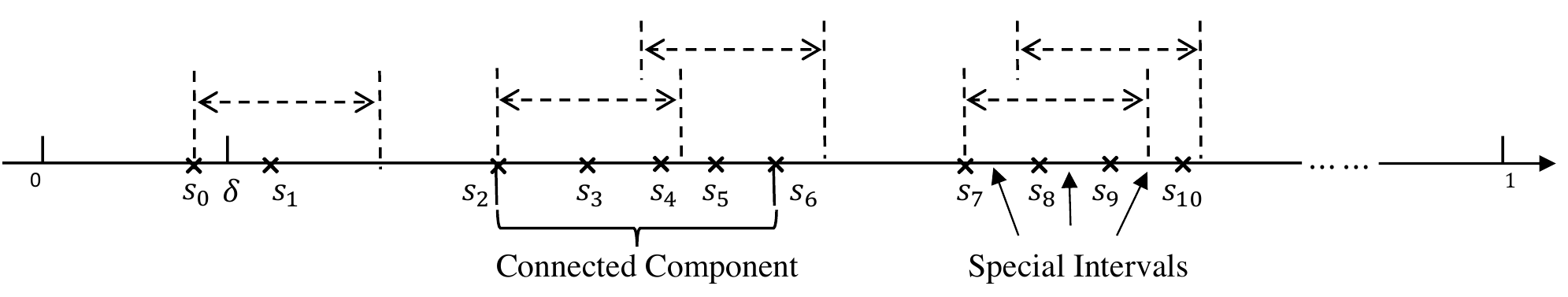}
   \caption{An illustration of the proof. $s_{1}-s_0\leq\epsilon$ and hence, $[s_0,s_{1}]$ is a special interval. There are three connected components: $[s_0,s_1]$, $[s_2,s_6]$, and $[s_7,s_{10}]$. The dashed lines with arrows represent a cover. }\label{lowerboundaa}
   \vspace{-0.25in}
 \end{figure}

By combining the results of the two cases, and using one more interval of length $\delta$ to cover the set $[0,\delta]$, we conclude that the $\delta$-cover number of $\overline{\cal I}(\overline{q})$, and therefore of ${\cal I}(\overline{q})$ as well, is at most $(K+3)/2$.
On the other hand, as long as we are dealing with an $(\epsilon,\delta,L)$-private strategy, this $\delta$-cover number is at least $L$. Thus $(K+3)/2\geq L$, or
$K\geq 2L-3$. 

Recall that the argument is being carried out for the case of particular $x_0$ and $y_0$ with the properties specified earlier. For such $x_0$ and $y_0$, we have at least $\log (\delta/\epsilon)$ queries in the set $[0,\delta)$ and at least $2L-3$ queries in the set $[\delta,1]$. 
Recall that we have introduced an additional artificial query at $x=1$, i.e., $q^{N+1}=1$, which is above and beyond the $N$ queries used by the strategy, and this query may be included in the $K$ special queries. For this reason, the lower bound on $N$ has to be decremented by 1, leading to the 
 lower bound in Theorem \ref{thm:1}. \qed

\section{Conclusions and Future Work}\label{sec:conclusion}
This paper studies an intrinsic privacy-complexity tradeoff faced by a {learner} in a sequential learning problem who tries to conceal her findings from an observant adversary. We use the notion of information set, the set of possible true values, to capture the information available to the adversary through the {learner}'s learning process, and focus on the coverability of the information set as the main metric for measuring a {learner} strategy's level of privacy. Our main result shows that to ensure privacy, i.e., so {that the resulting information set requires at least $L$ intervals of size $\delta$ to be fully covered}, it is necessary for the {learner} to employ at least $\log(\delta/\epsilon)+2L-4$ queries. We further provide a constructive {learner} strategy that achieves privacy with  $\log(1/L\epsilon)+2L$ queries. Together, the upper and lower bounds on the query complexity demonstrate that increasing the level of privacy, $L$, leads to a \emph{linear} additive increase in the {learner}'s query  complexity. 

{While mathematically valid and worst-case optimal, the Opportunistic Bisection Strategy could be potentially undesirable in applications where the learner aims for an extremely accurate estimate, i.e., $\epsilon\rightarrow 0$. Indeed, when $\epsilon$ is small, the chance that the guesses contain the true value becomes negligible. In practice, as long as the adversary is willing to tolerate a small chance of making an error, predicting according to the subsequent bisection phase would be much more beneficial: as $\epsilon$ goes to 0, the probability of accurately estimating $v^*$ approaches $1$ for the adversary. 
As was mentioned in Section \ref{sec:risk_averse_adversary}, this naturally motivates an average-case Bayesian variant that essentially focuses on covering most, rather than all, of the information set (Appendix B).
There, the OB strategy performs arbitrarily badly when $\epsilon$ is small enough, and the formulation warrants a new private strategy, which turns out to be the Replicated Bisection in Section \ref{sec:example_strategy}. We refer the readers to Appendix B for a much more elaborated discussion on the Bayesian variant. Relatedly,
as another direction of resolving the above practical issue, one could attempt to refine the information set to only contain those $x'$s whose probability of producing the observed sequence of queries is greater than some confidence level $\xi$. This is to be contrasted with only requiring a positive probability in the current formulation. More precisely, in Eq.~(\ref{eq:calQx}), one could let the requirement be $\mathbb{P}_{\phi}(Q_x=\bar{q})>\xi$, for some confidence level $\xi>0$. It would be interesting to see whether we can use the OB strategy as the base case to construct a class of strategies with respect to different confidence levels $\xi$.  }

{In addition,} there are several interesting extensions and  variations of the model that were left unaddressed. 
One may consider the binary query model in higher dimensions, {where $v^*\in [0,1)^d$.} A query in this setting will be a hyperplane in $\R^d$, where the response indicates whether the true value is to the right or to the left of the queried hyperplane. {We further investigate this extension in Appendix \ref{app:highdim}. 
Yet another interesting extension is to consider alternative response models, such as noisy binary responses, or real-valued responses that can convey a richer set of information.}
More broadly, there could be other interesting problem formulations for understanding the privacy implications in a number of sequential decision problems in learning theory, optimization, and decision theory. 
It is not difficult to see that standard algorithms, originally designed to optimize run-time or query complexity, often provide little or no protection for the {learner}'s privacy. Can we identify a universal procedure  to design sample-efficient private decision strategies? Is there a more general tradeoff between privacy and complexity in sequential decision making?  We are optimistic that there are many fruitful inquiries along these directions. 

\bibliographystyle{informs2014}
\bibliography{Private2017.bib}

\begin{APPENDIX}{}

\section{Coverability and the Adversary's Probability of Correct Estimation}
\label{sec:privacy_and_winning}

{In this section, we provide an alternative, probabilistic interpretation of the  definition of privacy in Definition \ref{def:secure_learner_strategy}. In particular, we show that $1/L$  can be interpreted as  a worst-case guarantee on the adversary's probability of correct detection.} 

 Recall the definition of $\calQ(x)$ as the set of all  possible query sequences when $v^*=x$ (cf. Eq.~\eqref{eq:calQx}), and let $\calQ = \cup_{x\in [0,1)} \calQ(x)$. We next define adversary 
 {estimators} $\hat{x}^a$ as random variables whose values are determined by the observed query sequence $\overline{q}$, together with an independent randomization seed.
\begin{definition}
\label{provider:win}
Fix $\delta>0$, $L\geq 2$, a {learner} strategy $\phi \in \Phi_N$, and a sequence of queries $\overline{q} \in \calQ$.  We say that  an adversary {estimator}, $\hat{x}^a$, is \emph{$(\delta,L)$-correct} with respect to $\overline{q}$,  if 
\begin{equation}\label{provider:successful}
\begin{split}
\mathbb{P}\Big(\left|\hat{x}^a(\overline{q})-x\right|\leq {\delta}/{2}\Big)>\frac{1}{L},  \quad {\forall \:x}\in\calI(\overline{q}). 
\end{split}
\end{equation}
where the probability is taken with respect to {any randomization} in the adversary's {estimator,} $\hat{x}^a$. 
\end{definition}  

In words, an adversary {estimator} is $(\delta,L)$-correct with respect to $\overline{q}$ if, as soon as the {learner} deploys the queries $\overline{q}$, the adversary will \emph{know} that the {resulting} estimate will incur  an error of at most $\delta/2$ with probability larger than $1/L$. In a sense, this means that $\overline{q}$ conceals the true value ``poorly.'' 

{Based on this probabilistic {definition of the correctness of adversary estimators,} we can then define a {new} notion of privacy for learner strategies, in a manner similar to Definition \ref{def:private_strategy}. To distinguish the two, we 
{use the term
``secure learner strategies'' for the new notion.}}

\begin{definition}[{Secure Learner Strategy}]
\label{def:secure_learner_strategy}
{Fix $\epsilon>0$, $\delta>0$, $L\geq 2$, with $L\in\mathbb{N}$. A {learner} strategy $\phi\in\Phi_N$ is $(\epsilon,\delta, L)$-secure if it satisfies the following: }
\begin{enumerate}
{\item Accuracy constraint: the {learner} estimate accurately recovers the true value, with probability one:
\begin{equation*}
\mathbb{P}\Big( \big|\hat{x}(x,Y)-x\big| \leq {\epsilon}/{2}\Big)=1, \quad \forall\: x\in[0,1), 
\end{equation*}
where the probability is measured with respect to the randomness in $Y$. }

{\item Privacy constraint: for every possible sequence of queries $\overline{q} \in \calQ$, there does not exist an adversary {estimator}, $\hat{x}^a$, that is $(\delta,L)$-correct with respect to $\overline{q}$. That is,  for every $\overline{q}\in\calQ$, there exists $x\in\calI(\overline{q})$ such that 
\begin{equation}
\begin{split}
\mathbb{P}\Big(\left|\hat{x}^a(\overline{q})-x\right|\leq {\delta}/{2}\Big)\leq\frac{1}{L}, 
\end{split}
\end{equation}
where the probability is taken with respect to {any randomization} in $\hat{x}^a$. } 
\end{enumerate}
\end{definition}

{The following proposition establishes an equivalence between the private learner strategies, defined in terms of the coverability of information set, and secure learner strategies, defined in terms of adversary's probability of correct estimation.}

\begin{proposition}
\label{prop:InfoSetProb}
{Fix $\delta>0$, $\epsilon>0$, $L\geq 2$, with $L\in\mathbb{N}$. A learner strategy $\phi\in\Phi$ is $(\epsilon,\delta,L)$-secure if and only if it is $(\epsilon,\delta,L)$-private. }
\end{proposition}

\bpf
{We  first establish the forward direction{: that} if $\phi$ is $(\epsilon,\delta,L)$-secure, then {$\phi$} is $(\epsilon,\delta,L)$-private. {We will actually establish the equivalent statement, that if
$\phi$ is not $(\epsilon,\delta,L)$-private, then it is not $(\epsilon,\delta,L)$-secure. For this,}
it suffices to show the claim that{,} for a sequence of queries $\overline{q} \in \calQ$, if {the $\delta$-cover number of the information set $\calI(\overline{q})$, $C_\delta\big(\calI(\overline{q})\big)$, is at most $L-1$} {(thus violating $(\epsilon,\delta,L)$-privacy),} then there exists an adversary {estimator} that is $(\delta,L)$-correct with respect to $\overline{q}$. }
{To show the claim,}
fix $\overline{q} \in \calQ$ such that $\calI(\overline{q})$ is {$(\delta,L-1)$-coverable}. Then, there exist $L-1$ intervals, $[a_1,b_1]$, $[a_2,b_2]$, $\dots$, $[a_{L-1},b_{L-1}]$, each of length $\delta$, that cover $\calI(\overline{q})$. Consider a randomized adversary {estimator} $\hat{x}^a$ that is distributed uniformly at random among the $L-1$ mid-points of the intervals. Then, with probability $1/(L-1)$, the {resulting} estimator $\hat{x}^a$ will lie in the same interval as the true value; 
{when this event occurs, and
since {the length of each one of the intervals is} at most $\delta$, the estimate will be at a distance of at most $\delta/2$ from the true value}, implying that
\begin{equation*}
\mathbb{P}\Big(\left|\hat{x}^a(\overline{q})-x\right|\leq {{\delta}}/{2} \Big)= \frac{1}{L-1}>\frac{1}{L},\qquad \forall x\in \calI(\overline{q}). 
\end{equation*}
This shows that $\hat{x}^a$ is $(\delta,L)$-correct given $\overline{q}$, which proves the claim.

{Conversely, we {will} now prove that if $\phi$ is $(\epsilon,\delta,L)$-private, then it is $(\epsilon,\delta,L)$-secure. It suffices to show the following claim: for {any} sequence of queries $\overline{q} \in \calQ$, if {$C_\delta\lt(\calI(\overline{q})\rt)$ is at least $L$}, then there does not exist an adversary {estimator} that is $(\delta,L)$-correct with respect to $\overline{q}$.} We will make use of the following lemma.

\begin{lemma}\label{lem:packing_variant} 
 Fix $\delta\in (0,1)$ and $L\geq 2$. Let $J $ be a subset of $[0,1)$ {such that the $\delta$-cover number of $J$, $C_\delta(J)$, is at least $L$.} Then, there exist points $\{x_1, x_2,\ldots, x_L\}$ in the closure of $J$ such that
 \begin{equation}
 \label{eq:lem_packing}
{|x_i-x_j| > \delta,  \quad \forall i\neq j.  }
 \end{equation}
\end{lemma}
\bpf {We will prove the lemma by constructing the set of $x_j$'s explicitly, with the aid of a ``helper'' sequence, $\{z_j\}$. In particular, we will construct $\{z_j\}$ {so} that the {sequence} of intervals $\{[z_j,z_j+\delta]\}$ forms a cover. The sequence $\{x_j\}$ will then be derived from a perturbed version of $\{z_j\}$, so that the constraint in Eq.~(\ref{eq:lem_packing}) can be satisfied. }

{Let $\overline{J}$ be the closure of $J$.  Consider the following procedure: 
\begin{enumerate}
\item Let $z_1=x_1 = \min \overline{J}$. 
\item For $i=2,3,\ldots$, consider the following recursive process of constructing $z_i$ and $x_i$: Let
\begin{equation*}
y_i \triangleq \min \{x\in \overline{J}: x \geq z_{i-1}+\delta\}.
\end{equation*}
Now, consider two scenarios:
\begin{enumerate}
	\item[I.] If $y_i>z_{i-1}+\delta$, then let $z_i=x_i=y_i$.
	\item[II.] If $y_i=z_{i-1}+\delta$, then check {whether} $y_i$ is a right endpoint of some interval in $\overline{J}$:
	\begin{enumerate}
			\item[(a)] if there exist{s} $\lambda_i>0$ small enough such that $[y_i,y_i+\lambda_i]\subset\overline{J}$, then let $z_i=y_i$ and $x_i=y_i+\lambda_i'$, where $0<\lambda_i'<\lambda_i$ and $\lambda_i'$ is sufficiently small.
		\item[(b)] Otherwise, 
		{if such a $\lambda_i$ does not exist, let}
		\begin{equation*}
z_i = x_i=\min \{x\in \overline{J}: x > z_{i-1}+\delta\}. 
\end{equation*} 
	\end{enumerate}
\end{enumerate}
\end{enumerate}
The procedure terminates at some step $T$ when 
$(z_T+\delta, 1) \cap \overline{J} = \emptyset$. Note that by construction, all $z_i$'s and $x_i$'s belong to the closure of $J$. Furthermore, the intervals 
\begin{equation*}
W_i := [z_i, z_i+\delta], \quad i=1, 2, \ldots, T, 
\end{equation*}
form a cover of $\overline{J}$. Since {$C_\delta(J)\geq L$ by assumption}, it follows that we must have $T\geq L$. Finally, it is easy to verify that the points $\{x_1, x_2, \ldots, x_L\}$ satisfy the conditions outlined in the lemma (Eq~(\ref{eq:lem_packing})). In particular, this is guaranteed by choosing $\lambda_i'$ to be sufficiently small and $\lambda_i'<\lambda_j'$ for $i<j$, when Case II-(a) above occurs. This completes the proof. }\qed

{Fix {an} adversary {estimator} $\hat{x}^a$ and {some} $\overline{q}\in\calQ$ such that {$C_\delta\big(\calI(\overline{q})\big)\geq L$.}
 Apply Lemma \ref{lem:packing_variant} with $J = \calI( \overline{q})$, and  let $\{x_1, x_2, \ldots, x_L\}$ be as defined in the lemma. Because the $x_i$'s belong to the closure of $\calI(\overline{q})$ and $|x_i-x_j|>\delta$ for any $i\neq j$, by slightly perturbing them, we can obtain a set of points $\{\tilde{x}_1,\tilde{x}_2, \ldots, \tilde{x}_T\} \subset \calI(\overline{q})$, 
such that 
\begin{equation*}
|\tilde{x}_i - \tilde{x}_j |> \delta, \quad \forall i\neq j,
\end{equation*}
still hold{s}.} Define intervals
 \begin{equation*}
 U_i := [\tilde{x}_i-\delta/2,\: \tilde{x}_i+\delta/2], \quad i = 1, 2, \ldots, L. 
 \end{equation*}
Since the distance between any two distinct $\tilde{x}_i$'s is greater than $\delta$, we know that the  intervals $U_i$ are disjoint, which implies that 
{at least one of these $L$ intervals will have probability {of containing $\hat{x}^a(\bar{q})$} less than or equal to $1/L$. In particular,} 
there exists $i^* \in \{1, 2, \ldots, L\}$ such that 
\begin{equation*}
\pb\Big(|\what{x}^a(\overline{q})-\tilde{x}_{i^*}| \leq  \delta/2\Big) {=} \pb(\what{x}^a( \overline{q}) \in U_{i^*} ) \leq  1/L. 
\end{equation*}
Since $\tilde{x}_{i^*} \in \calI(\overline{q})$ by construction, we conclude that the adversary estimate $\hat{x}^a$ is not $(\delta, L)$-correct with respect to $ \overline{q}$. This completes the proof of the claim and hence, the converse direction of the proposition.
 \qed

\section{Bayesian Private {Learning} Model}
\label{app:baysianprivate}
The Private Sequential Learning model we studied in this paper assumes that neither the {learner} nor the adversary has any prior information {on} the true value $v^*$, and {that} they {can obtain information only gradually, through queries}. In this section, we discuss a Bayesian variant of the model,  where the true value $v^*$ is generated according to a prior distribution,  known to both parties. In particular, we assume that $v^*$ is distributed according to a distribution $P_{v^*}$, where, for simplicity, we assume that the support of $P_{v^*}$ is equal to $[0,1)$. 

Naturally, we allow the learner strategy, defined in Section \ref{sec:user_strategy}, to depend on $P_{v^*}$. 
In addition, since the adversary aims to produce an estimate $\hat{x}^a$ that is close to the true value, we define an adversary strategy  to be a function {$\psi$} that maps the adversary's available information (i.e., the prior distribution $P_{v^*}$, the {learner} strategy $\phi$, and the sequence of observed queries $\overline{q}$) to a probability distribution over $[0,1)$, or, equivalently, to a 
random variable, $\hat{x}^a$, that takes values in $[0,1)$. Denote by $\Psi$ the set of all such functions, i,e., the set of all adversary strategies.

Note that, since the true value in the Bayesian Private Learning model admits a prior distribution, instead of using the information set, it is sufficient for the adversary to keep track of the  posterior distribution of $v^*$, given the {learner}'s queries. The Bayesian formulation also allows us to measure the probability that the adversary is able to provide an estimate of the true value that is within a given error tolerance. This leads to the following definition of {Bayesian} private {learner} strategies. 
\begin{definition} \label{def:bayes_private_strategy}
{Fix $\epsilon>0, \delta>0$, 
{an integer $L\geq 2$,} and a prior distribution $P_{v^*}$. A {learner} strategy $\phi\in\Phi_N$ is $(\epsilon,\delta, L)$-{B-private} if it satisfies the following: 
\begin{enumerate}
\item Accuracy constraint: the strategy accurately recovers $v^*$, with probability one: 
\begin{equation*}
\mathbb{P}\left(\big|\hat{x}(v^*,Y)-v^*\big|\leq\frac{\epsilon}{2}\right)=1,
\end{equation*}
where the probability is taken with respect to the randomness in $v^*$ and $Y$. 
	\item Privacy constraint: under this learner strategy, and for every adversary strategy $\psi\in\Psi$, we have
	\begin{equation}
\mathbb{P}\big(|\hat{x}^a-v^*|\leq \delta/2\big) \leq \frac{1}{L},
	 \end{equation} 
\end{enumerate}}
where the probability is taken with respect to the randomness in $v^*$, $Y$, and $\hat{x}^a$. 
\end{definition}
Notice the resemblance of the above definition with Definition \ref{def:private_strategy}. The parameters $\epsilon$ and $\delta$ have the same meaning as in the original model, while $L$ mirrors the role of $L$ in $(\delta,L)$-coverability{,} but now has a more concrete interpretation in terms of the adversary's probability of error.

Fix a prior distribution $P_{v^*}$. Denote by $N_{{B}}^*(\epsilon,\delta,L)$ the minimum number of queries needed for there to exist an $(\epsilon,\delta,L)$-{B-private} {learner} strategy: 
\begin{equation*}
N_{{B}}^*(\epsilon,\delta,L)=\min\big\{N\in\mathbb{N}: \exists\:\phi\in\Phi_N \text{ s.t. $\phi$ is $(\epsilon,\delta,L)$-{B-private}}\big\}. 
\end{equation*}
Similar to the original model, we would like to obtain lower and upper bounds on $N_{{B}}^*(\epsilon,\delta,L)$.

\subsection{{Comparing the two formulations}}
{In single-player optimization, Bayesian formulations are always less demanding than worst-case formulations. However, this is not the case in game-theoretic situations, and an automatic comparison between $N^*(\epsilon,\delta,L)$ and $N_{{B}}^*(\epsilon,\delta,L)$ is not available. This is because the Bayesian formulation can, in principle, work in the favor of either player, as discussed below:}
\begin{enumerate}
\item[(i)]
{Under the worst-case formulation, once an information set is determined, and in order to break the learner's privacy, the adversary must be able to cover (with  intervals of length $\delta$) the \emph{entire} information set. In contrast, in  the Bayesian formulation, the adversary can break the learner's privacy by covering \emph{most} of the information set. Since the Bayesian formulation seems to make the adversary's objective easier to achieve, one might expect that the situation has become more demanding for the learner, resulting in higher query complexity.}
\item[(ii)]
{Under the worst-case formulation, Part 2 of Definition \ref{def:private_strategy} involves a privacy requirement that holds ``for every $x\in[0,1)$.'' In a Bayesian formulation, this is essentially replaced by a requirement that holds ``on the average,'' over $x$. In principle, this is an easier requirement for the learner, and might work in the direction of lower query complexity under a Bayesian formulation.}
\end{enumerate}

{For a concrete illustration of the difference between the two formulations, consider the Opportunistic Bisection strategy in Section \ref{sec:upperbound}, which was private under the worst-case formulation. 
Under that strategy, privacy was ensured by having many small intervals in the information set (the ``guesses'') that the adversary could not ignore. On the other hand, under the Bayesian formulation, 
and as long as $\epsilon$ is small enough, the probability that these small intervals contain $v^*$ is negligible, and privacy is {completely} lost in the Bayesian setting.} {Specifically, suppose that the prior distribution of $v^*$ is uniform over $[0,1)$, and consider an adversary strategy that always uses the last query of Phase 2 of the Opportunistic Bisection Strategy as her estimator, i.e., the last query of the local bisection search phase. Recall the regime of interest, $2\epsilon< \delta\leq 1/L$. With probability $1-L\epsilon$, the $L$ ``guesses'' will not contain $v^*$, and Phase 2 would be a truthful bisection search, under which the above adversary strategy would succeed. That is, the probability that the adversary recovers $v^*$ with error at most $\delta/2$ is at least $1-L\epsilon$. The above arguments demonstrate that when $\epsilon$ is too small, privacy cannot be guaranteed under the Bayesian formulation.
}

\subsection{Results}

 For the case where $P_{v^*}$ is a uniform distribution over $[0,1)$, we can obtain the following result by adapting the proof for the original model. 
\begin{proposition} 
\label{thm:bayes}
Fix $\epsilon>0$, $\delta>0$, and a positive integer $L\geq 2$, such that $2\epsilon< \delta\leq{1}/{L}$. Suppose that the prior distribution $P_{v^*}$ is uniform over $[0,1)$.  
Then, 
\begin{equation*} 
\log\frac{1}{\epsilon} \leq {N_B^*(\epsilon,\delta,L)}\leq L\log\frac{1}{L\epsilon}+L-1.
\end{equation*}
\end{proposition} 

{The lower bound, $\log(1/\epsilon)$, follows directly from the accuracy constraint (see Section \ref{sec:lowerbound}). The upper bound is achieved by the Replicated Bisection strategy in Section \ref{sec:example_strategy}, with $L$ replications. Because the Replicated Bisection strategy creates $L$ identical copies of query patterns across $L$ sub-intervals, the symmetry ensures that the posterior distribution of $v^*$ will be evenly spread across all sub-intervals, forcing the adversary's probability of correct detection to be at most $1/L$.}

{Note that, in contrast to the bounds in Theorem \ref{thm:1}, the leading terms in the upper and lower bounds in Proposition \ref{thm:bayes} differ by a multiplicative factor of $L$, and closing this gap is non-trivial. In particular, the approach used in this paper to prove the lower bound 
is highly dependent on the structure of the information set and is unlikely to apply.}\footnote{{In a recent paper that follows up on our work, \cite{kuangNIPS2018} shows that the upper bound in Proposition \ref{thm:bayes} is asymptotically tight up to the first order, in the limit as $\epsilon\to 0$, while $L$ and $\delta$ remain fixed. The proof in \cite{kuangNIPS2018} employs an information-theoretic argument that is {very different from the line of analysis in this paper.}} }

\section{High-dimensional Private Sequential Learning}
\label{app:highdim}
{
In this section, we provide a brief discussion {of a} high-dimensional variant of the Private Sequential Learning model that allows us to apply the insights from the original model {to} a more general setting.  
}

{
Consider a learner who {wants} to identify a true value $v^*$ situated in a $d$-dimensional unit cube, $[0,1)^d$.  A query is now a hyperplane in $\R^d$, where the response indicates whether the true value is to the right or to the left of the queried hyperplane. Formally, a query {is specified by some} $q=(q^{(1)},q^{(2)},\dots,q^{(d)},c)\in\R^{d+1}$, and {the corresponding} response  {is} $r=\mathbb{I}(\langle(q^{(1)},\dots,q^{(d)}),v^*\rangle\leq c)$. The distance metric for measuring estimation error{s} is the $l_\infty$ norm.
In this setting, the adversary 
{aims to cover the information set using less than $L$}  
hypercubes of edge length $\delta$ (and hence volume $\delta^d$). 
}

{Let us focus on
the regime where $2\epsilon<\delta\leq 1/L^{1/d}$, 
{and assume, for the sake of this discussion, that $L^{1/d}$ is an integer.} For {an} upper bound, it is not hard to see that we can extend the Opportunistic Bisection strategy to this case in a straightforward manner. In particular, we {can use} the following two-phase strategy:}

{{\emph{Phase 1 - Opportunistic Guesses.}
During the first phase, we first partition the space into $L$ equal-sized ``large'' hypercubes, each with volume $1/L$, {so that each side of a hypercube has length $L^{-1/d}$.} This can be achieved {by} submitting $L^{1/d}$ queries for each dimension. For example, the queries for the first dimension are $\{(1,0,\dots,0,(i-1)L^{-1/d})\}_{i=1}^{L^{1/d}}$ and the queries for the second dimension are $\{(0,1,\dots,0,(i-1)L^{-1/d})\}_{i=1}^{L^{1/d}}$. To achieve successful obfuscation, so that the adversary is not able to cover the information set with $L-1$ hypercubes, the second step of this phase is to construct {a smaller sub-cube,  with edge length $\epsilon$ inside each one of the $L$ hypercubes,} to serve as the plausible ``guesses''. Again, this is achieved by another $L^{1/d}$ queries for each dimension. For instance, the queries for the first dimension are $\{(1,0,\dots,0,(i-1)L^{-1/d} + \epsilon)\}_{i=1}^{L^{1/d}}$ and the queries for the second dimension are $\{(0,1,\dots,0,(i-1)L^{-1/d} + \epsilon)\}_{i=1}^{L^{1/d}}$. }}

{{\emph{Phase 2 - Local Bisection Search.}
{After the} guesses {are} created in Phase 1, the learner {then} performs a bisection search inside one of the $L$ large hypercubes (i.e., $d$ bisection search{es} along each dimension). Depending on whether the guesses contains $v^*$ or not, this bisection search is either truthful or fictitious, similar to the original policy. }}

{Using the same argument as in  the analysis for the original Opportunistic Bisection strategy, it is not difficult to verify that the above strategy is private and achieves a query complexity of $d\log(1/(L^{1/d}\epsilon))+2dL^{1/d}$. For the lower bound, it is evident that the agent needs at least $d\log(1/\epsilon)$ queries to {staisfy the accuracy constraint,} even {in the absence of a}  privacy constraint. It appears challenging however to obtain a stronger lower bound that depends on the dimension in a meaningful way. The argument that we employ in the proof of the current lower bound does not generalize easily {to} higher dimensions. 
}




\end{APPENDIX}

\end{document}